\documentclass[11pt,a4paper]{article}
\usepackage[hyperref]{emnlp2020}
\usepackage{times}
\usepackage{latexsym}
\pdfoutput=1

\usepackage{microtype}

\usepackage{url}
\usepackage{amsmath}
\usepackage{amssymb}
\usepackage{multirow}
\usepackage{tikz-dependency}
\usepackage{caption}
\usepackage{subcaption}
\usepackage{cleveref}

\usepackage{enumitem}

\tikzset{every picture/.style={line width=0.4pt}} 

\aclfinalcopy 
\def\aclpaperid{61} 

\title{Recurrent babbling:\\evaluating the acquisition of grammar from limited input data}

\author{Ludovica Pannitto\\
  CIMeC\\
  University of Trento\\
  \texttt{ludovica.pannitto@unitn.it} \\\And
  Aurélie Herbelot\\
  CIMeC/DISI\\
  University of Trento\\
  \texttt{aurelie.herbelot@unitn.it} \\}

\date{}

\begin{document}
\maketitle
\begin{abstract}
Recurrent Neural Networks (RNNs) have been shown to capture various aspects of syntax from raw linguistic input.
In most previous experiments, however, learning happens over unrealistic corpora, which do not reflect the type and amount of data a child would be exposed to. This paper remedies this state of affairs by training a Long Short-Term Memory network (LSTM) over a realistically sized subset of child-directed input. The behaviour of the network is analysed over time using a novel methodology which consists in quantifying the level of grammatical abstraction in the model's generated output (its `babbling'), compared to the language it has been exposed to. We show that the LSTM indeed abstracts new structures as learning proceeds.
\end{abstract}

\section{Do RNNs learn grammar?}
\label{sec:rnn}

Artificial Neural Networks, and Long Short-Term Memory Networks more specifically, have consistently demonstrated great capabilities in the area of language modeling. In addition to generating credible surface patterns, they show excellent performances when tested on very specific grammatical abilities~\cite{gulordava2018colorless,lakretz2019emergence}, without requiring any prior bias towards the syntactic structure of natural languages. 

From a theoretical point of view, these results seem to contradict the well-known argument of the \textit{poverty of the stimulus}~\cite{chomksy1959review,chomksy1968language} and raise questions about the \textit{continuity hypothesis} in language acquisition~\cite{lust1999universal,crain2001nature}. At the same time, a number of results give a much more mitigated view of RNNs' abstraction capabilities \citep{marvin2018targeted,chowdhury2018rnn}. It thus remains unclear how and to what extent grammatical abilities emerge in artificial language models, and how this knowledge is encoded in their representations -- especially when considering notions such as \textit{productivity} and \textit{compositionality}~\cite{baroni2020linguistic}, which are recognised as defining traits of natural languages.

This paper proposes that the evaluation of RNN grammars should be widened to include the effect of the type of input data fed to the network, as well as the theoretical paradigm used to analyse its output. We specifically remark that much of the discussion concerning language modeling remains influenced by the mainstream \textit{generativist} approach, which posits a sharp distinction between syntax and the lexicon. Our own approach will be to depart from this account by testing the grammatical abilities of an RNN in a \textit{usage-based} perspective. Specifically, we ask what kind of structures are abstracted and used productively by the network, and how the abstraction process takes place over time. 

In contrast with previous models: (i) we train a vanilla char-LSTM on a more realistic variety and amount of data, focusing on a limited amount of child-directed language; (ii) we do not rely on extrinsic evaluations or downstream tasks, instead we introduce a methodology to evaluate how the distribution of grammatical items, over time, comes to approximate the one in the input, through a continuous process and (iii) we tentatively explore the interaction between meaning representations and the abstraction abilities of the network, blurring the distinction between lexicon and syntax, in a way more akin to Construction Grammar (CxG, \citealp{fillmore1988mechanisms,goldberg1995constructions,kay1999grammatical}). Our evaluation focuses on the network's generated output (its `babbling'), asking to what extent the system simulates the type of grammatical abstraction observed in human children. The study is conducted on English.

In what follows, we review related work (\S~\ref{sec:relwork}), we then formulate the question of grammar modelling in a broader theoretical framework (\S~\ref{sec:framework}) involving three parameters: the type of acquisition mechanism under study, the nature of the input data, and the representational paradigm adopted for the analysis. We configure this broad framework with particular choices of parameters and implement it in \S~\ref{sec:data}, \ref{sec:q1} and \ref{sec:q2}. We provide two analyses of the distributional properties of the network's `babbling', discussed in \S~\ref{sec:discussion}.

\section{Related Work}
\label{sec:relwork}

A considerable amount of literature has investigated the ability of ANNs to acquire grammar, and the list we present here is by no means exhaustive. The analysis of the syntactic abilities of LSTMs~\cite{hochreiter1997long} and ANN-based language models dates back quite a few years~\cite{mcclelland1992can,lewis2001learnability}. Recent contributions have followed a general tendency to analyze the inner-workings of networks, and the specific type of knowledge they acquire~\cite{alishahi2019analyzing,linzen2020syntactic}. For instance, \citet{linzen2016assessing} show how a network acquires \textit{abstract} information about number agreement, albeit in a supervised setting. The same study is expanded in \citet{gulordava2018colorless}, which shows how a language modeling task is enough for a network to predict long-distance number agreement, both on semantically sound and nonsensical sentences. The authors conclude that ``LM-trained RNNs can construct abstract grammatical representations'', but their model is trained on a rather consequent amount of data (90M tokens) from a rather peculiar distribution (a Wikipedia snapshot). Similarly, it has been shown that LSTMs ~\cite{mccoy2018revisiting,wilcox2018rnn} can learn tricky syntactic rules like the English auxiliary inversion and filler-gap dependencies, although, in later work, \citet{mccoy2020does} find that only models with an explicit inductive bias~\cite{shen2018ordered} learn to generalize the \textsc{move-main} rule with respect to auxiliary inversion.
\citet{marvin2018targeted} show instead poor performance of RNNs in grammaticality evaluation, due to their sensitivity to the specific lexical items encountered during training, a limitation that, they say, ``would not be expected if its syntactic representations were fully abstract''. \citet{chowdhury2018rnn} similarly state that their model ``is sensitive to linguistic processing factors and probably ultimately unable to induce a more abstract notion of grammaticality''.

Moreover, despite the fact that the model of \citet{gulordava2018colorless} is tested on four languages, the most promising results may not be generalizable to languages showing different surface patterns from English. \citet{ravfogel2018can} fail to replicate \citet{gulordava2018colorless}'s results on Basque, and \citet{davis-van-schijndel-2020-recurrent}, after testing the network on relative clause attachment cases in English and Spanish, conjecture that the associative (non-linguistic) bias of RNNs overlaps with English syntactic structure but represents an obstacle to learn attachment rules for Spanish.

Other puzzling results concern the relation of perplexity to syntactic performance~\cite{warstadt2019blimp,hu2020systematic}: having evaluated their models on 34 benchmarks, \citet{hu2020systematic} conclude with a call for a wider variety of syntactic phenomena to test on.
Further studies have shown that networks carrying explicit inductive bias perform better than vanilla LSTMs. In a recent paper, \citet{lepori2020representations} show that a constituency-based network generalizes more robustly than a dependency-based one, and that both outperform a more basic BiLSTM.
Lastly, we mention the study carried out by \citet{kuncoro-etal-2018-lstms} who perform their study using a character-based LSTM -- a choice we will follow in this work.

A very similar scientific discussion, which we won't report in depth here, is blooming around Transformer-based language models~\cite{tran2018importance,goldberg2019explain,bacon2019does,jawahar2019does,lin2019open}, leading to similar contrasting results.

Finally, a separate line of work focuses on a more indirect test of the information encoded in the internal representation, assessing which aspects of the original syntactic structure can be reconstructed through diagnostic classifiers~\cite{adi2017fine,giulianelli2018under,hewitt2019structural,tenney2019you}.

In summary, a clear trend has not yet emerged~\cite{linzen2020syntactic}. All the models we cited, however, seem to idealize syntactic structure as a separate and more abstract ability from the knowledge of statistical regularities or lexical co-occurrences. This perspective may reflect a belief in a sharp distinction between the \textit{lexicon} and \textit{compositional rules}. That is, ANNs are expected to gain \textit{abstract grammatical abilities} through compositional generalization, where compositionality is understood as the ability to produce an unbounded number of sentences by means of a set of algebraic rules~\cite{baroni2020linguistic}. In contrast with this approach, usage-based models encourage us to adopt a different perspective, and to analyze LSTMs' grammatical abilities with respect to the kind of representations (more in \S\ref{sec:framework-rep}) posited by theories such as Construction Grammar (CxG, \citealp{fillmore1988mechanisms,goldberg1995constructions,kay1999grammatical}).

\section{Framework}
\label{sec:framework}

In essence, the question of language acquisition asks how much language ($\Lambda$) can be learned with a certain level of computational complexity ($C$) by being exposed to a certain type of data ($I$). The corresponding formalization, $a: C \times I \mapsto \Lambda$, describes both human and artificial acquisition processes, and its components have been central in the linguistic debate. Below, we will discuss each term ($C$, $I$ and $\Lambda$) in further detail.

\subsection{Computational complexity of the acquisition mechanism ($C$)}

Our aim is to test how much grammatical structure can be induced from linguistic input through a pattern-finding mechanism such as that provided by ANNs. Therefore, we fix the level of computational complexity to a vanilla, character-based LSTM, which we train exploring different sources of input in a specific range $\{I_i\}$, selected based on their complexity level. We then use the trained model to generate some amount of text (to \textit{babble}), to explore the structure of the produced output $\ell \in \Lambda$, mainly with respect to productivity.

\begin{equation}
\left( LSTM, I_i \right)  \xrightarrow{a} \ell_i
\end{equation}

\vspace{-1mm}
Our choice of model has consequences from a theoretical point of view. Different stances have been taken about how much has to be \textit{hard-coded} or \textit{innate} in order for language acquisition to happen: while formal innatist theories have always posited the need for a specialized and innate ability, a dedicated \textit{device} for language learning~\cite{chomsky1981lectures,chomsky1995minimalist,hauser2002faculty}, cognitive theories have argued for a more systemic vision, showing how general purpose memory and cognitive mechanisms can account for the emergence of linguistic abilities~\cite{tomasello2003constructing,goldberg2006constructions,christiansen2016creating,cornish2017sequence,lewkowicz2018learning}.

LSTMs, under this perspective, can be seen as a domain-general attention and memory mechanism, without any explicitly hard-coded grammatical knowledge. They have been applied, without substantial modifications, to a variety of tasks, ranging from time series prediction to object co-segmentation, and encompassing grammar learning as well. On the continuum between specialized devices and general purpose associative mechanisms, LSTMs place themselves on the latter side, with their recurrent structure seeming to be crucial in the linguistic abstraction process~\cite{tran2018importance}.

\subsection{Structure and role of the input ($I$)}

Because of the traditional sharp distinction between \textit{competence} and \textit{performance}, the role of the input and the linguistic environment has been minimized by theories in the realm of Universal Grammar (UG).
Usage-based theories, on the other hand, have granted the input a central role to the end of explaining why language is structured as it is~\cite{fillmore1988mechanisms,kay1999grammatical,hoffmann2013constructions,christiansen2016creating,goldberg2019explain}: one of the striking points to make here is that in the usage-based framework the acquisition problem is framed as an incremental process. Acquiring language essentially entails learning how to process the linguistic input in an error-driven procedure, where full linguistic creativity and productivity are acquired gradually by speakers \cite{bannard2009modeling}, building up on knowledge about specific items and restricted abstractions.

In this sense, the specific features of the language on which ANNs are trained cannot be overlooked when it comes to describing their acquired grammatical abilities.
Compared to what a child is exposed to during the most crucial months of language acquisition, ANNs are trained on an input that is often unrealistic in size: the LSTM introduced in~\citet{gulordava2018colorless} is for example exposed to 90M tokens, and sees them multiple times over training.
It is hard to come up with a precise estimate of the amount of language children are exposed to during the years of acquisition, as the variation depends on a huge number of factors including the socio-economic environment~\cite{bee1969social} or the societal organization~\cite{cristia2019child}.
\citet{hart1995meaningful}, in a seminal work, estimate that, by the age of 3, welfare children have heard about 10 millions words while the average working-class child has heard around 30 millions. Finally, the domain of the data also matters: child-directed language is characterized by specific features~\cite{matthews2010children} that are not present in the most widely used corpora.\footnote{Specifically, those that contain data harvested from the web such as Wikipedia or UKWaC.}

\subsection{Shape and features of the generated language ($\Lambda$)}
\label{sec:framework-rep}

Any analysis of the language $\Lambda$ generated by a learner implies the availability of a \textit{representation}. Much has been written on the respective benefits of various representations of linguistic structures: the exact nature of their shape and content is the ultimate conundrum of linguistic theory. Of course, this paper is not the place to review the wide variations that exists among theories, so we will just limit ourselves to motivate our choice with respect to the broader theoretical framework.

Constituency-based representations have been prevalent in the description of natural language syntax, becoming primarily associated with derivational theories. Due to the Fregean view of compositionality, they have also become the natural building blocks for meaning composition.
Dependency representations have, on the other hand, re-gained popularity over constituency representations in the last decades, showing desirable properties from a computational perspective (they adapt to a wider array of languages, representing ill-formed sentences results easier and the output is more easily incorporated in semantic graphs) while taking a more functional approach to language description, more in line with cognition oriented-approaches.

In order to represent the features of $\ell \in \Lambda$, we choose a representation which makes the least possible assumptions on the acquisition process and on the content of the generated language, and is at the same time flexible and computationally tractable.
We therefore rely on dependency representations, more specifically the universal dependencies framework~\cite{nivre2020universal}, from which we extract subtrees called \textit{catenae} \cite{osborne2012catenae}. As we will see below, the notion of catena is more flexible than that of constituent, and allows us to describe a larger set of generalizations.

Generally speaking, CxG approaches seem to lack a shared representational framework\footnote{an exception should be made for the formalisms derived from the FrameNet project (\url{https://framenet.icsi.berkeley.edu/})}, relying on box diagrams or Attribute-Value Matrices to describe the traits of the fragments they study. The structures introduced by~\citet{osborne2006beyond} are characterized instead as fundamental meaning-bearing units~\cite{osborne2012constructions}, in line with the theoretical tenets of CxGs, thus being ideal candidates for the lexicon (or \textit{`Constructicon'}) postulated in such theories: catenae have in fact been applied in the description of construction-like structures~\cite{osborne2012constructions,dunn2017learnability} and allow for the representation of non-adjacent structures while encompassing the notion of constituent as well~\cite{osborne2006beyond,osborne2018tests}.

A catena is defined as ``a word, or a combination of words which is continuous with respect to dominance''~\cite{osborne2012catenae}: given a dependency tree, this definition selects a broader set of elements than the definition of constituent\footnote{which can be seen as a subtype of catena as ``A catena that consists of a word plus all the words that that word dominates''}. 
Unlike constituents, catenae can include both contiguous and non contiguous words. They however capture something more refined than generic subsets of sentence items, as the elements are grouped depending on the syntactic links holding in the sentence.

From a graph-theory perspective, catenae form subtrees (i.e., subsets of nodes and edges that constitute a tree themselves) of the original tree.

\begin{figure*}[t]
	\centering
	\begin{subfigure}[b]{0.25\textwidth}
		\centering
		\begin{dependency}[theme = simple]
			\begin{deptext}[column sep=0.3em]
				A \& B \& C \& D \& E \& F \& G \\
			\end{deptext}
			\deproot[edge unit distance = 2ex]{1}{ROOT}
			\depedge{1}{2}{}
			\depedge{1}{3}{}
			\depedge{1}{4}{}
			\depedge{1}{5}{}
			\depedge{1}{6}{}
			\depedge{1}{7}{}
		\end{dependency}
		\caption{The case of a flat structure, where all nodes are linked to the \textit{root}: from a tree like this we can extract $2^6-1$ catenae, each one containing $A$ plus a subset of its children nodes.}
		\label{fig:dep_flat}
	\end{subfigure}
	\hfill
	\begin{subfigure}[b]{0.25\textwidth}
		\centering
		\begin{dependency}[theme = simple]
			\begin{deptext}[column sep=0.3em]
				A \& B \& C \& D \& E \& F \& G \\
			\end{deptext}
			\deproot[edge unit distance = 1ex]{1}{ROOT}
			\depedge{1}{2}{}
			\depedge{2}{3}{}
			\depedge{3}{4}{}
			\depedge{4}{5}{}
			\depedge{5}{6}{}
			\depedge{6}{7}{}
		\end{dependency}
		\caption{The case where nodes are arranged in a full dependency chain: here the number of catenae corresponds to the number of substrings that could be extracted from the linear signal, that is $20$.}
		\label{fig:dep_chain}
	\end{subfigure}
	\hfill 
		\begin{subfigure}[b]{0.4\textwidth}
		\centering
		\begin{dependency}[theme = simple]
			\begin{deptext}[column sep=0.3em]
				A \& B \& C \& D \& E \& F \& G \\
			\end{deptext}
			\deproot[edge unit distance = 1ex]{4}{ROOT}
			\depedge{4}{3}{}
			\depedge{4}{5}{}
			\depedge{3}{1}{}
			\depedge{3}{2}{}
			\depedge{5}{6}{}
			\depedge{5}{7}{}
		\end{dependency}
		\caption{The case of a hierarchical structure, typically what we would find in linguistic trees, where the counts are less trivial to make. In particular, for each node we find that the number of catenae \textit{rooted} in that node can be estimated depending on the number of catenae \textit{rooted} in his children nodes, and depends therefore on the specific structure of the tree.}
		\label{fig:dep_hier}
	\end{subfigure}
	\caption{}
	\label{fig:trees}
\end{figure*}

Let's consider for example the structures represented in Figures~\ref{fig:trees}\subref{fig:dep_flat}, \ref{fig:trees}\subref{fig:dep_chain} and \ref{fig:trees}\subref{fig:dep_hier}: the same elements (nodes $A$ to $G$) are arranged differently in the structure of dependency tree, and this leads to a different number and composition of catenae.

\begin{figure}[t]
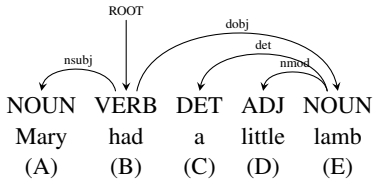

	\centering
	\begin{footnotesize}
	\begin{dependency}[theme = simple]
		\begin{deptext}[column sep=0.3em, row sep=.1ex]
			NOUN \& VERB \& DET \& ADJ  \& NOUN \\
			Mary \& had  \& a  \& little \& lamb  \\
			 (A)  \& (B)  \& (C) \& (D)   \& (E) \\   
		\end{deptext}
		\deproot[edge unit distance = 2ex]{2}{ROOT}
		\depedge{2}{1}{nsubj}
		\depedge{2}{5}{dobj}
		\depedge{5}{3}{det}
		\depedge{5}{4}{nmod}
	\end{dependency}
	\caption{The dependency representation of the sentence \textit{Mary had a little lamb}, annotated with morpho-syntactic and syntactic information.}
	\label{fig:mary-had-a-little-lamb}
	\end{footnotesize}
\end{figure}

\begin{table}[t]
\begin{footnotesize}
\begin{tabular}{p{0.12\textwidth}|p{0.3\textwidth}}
Strings & A, AB, ABC, ... B, BC, ...E \\ \hline
Catenae & A, B, C, D, E, AB, ABCE, ABDE, ABCDE, ABE, BCE,BDE, BE, CE, DE, CDE \\ \hline
Constituents & A, ABCDE, C, D, CDE
\end{tabular}
\caption{Possible structures that can be extracted from the dependency tree in Figure~\ref{fig:mary-had-a-little-lamb}}
\label{tab:mary-table}
\end{footnotesize}
\end{table}

As a concrete example, Figure~\ref{fig:mary-had-a-little-lamb} represents a dependency tree, and Table~\ref{tab:mary-table} the structures that can be extracted from it: considering the lexical level, we can extract \textit{Mary had lamb}, \textit{had a lamb}, \textit{a little lamb} as catenae. As the morpho-syntactic and syntactic levels are available, however, we can also extract partially filled structures as \textit{Mary had NOUN}, \textit{nsubj VERB dobj} and so on. 

Of interest for our analysis, CxG argues that grammar items above the lexical level bear meaning themselves, and that this emerges from patterns of usage. 
According to \citet{goldberg2006constructions}, for example, the meaning of the ditransitive pattern \textit{Sbj V Obj Obj2}, and thus its productivity, emerges from its strong association with \textit{give} in child-directed speech: part of the meaning of \textit{give} remains attached to the construction. A natural, and promising~\cite{rambelli2019distributional}, solution to represent the semantics of catenae is given by Distributional Semantics~\cite{harris1954distributional}, where each element of the \textit{`Constructicon'} is implicitly described in terms of its context of use~\cite{erk2012vector,lenci2018distributional}. We will see in \S\ref{sec:q2} how we can use such distributional representations to investigate the level of abstraction of our network's babbling.

\section{Data and language modelling}
\label{sec:data}

\subsection{Corpus}
\label{sec:corpus}

Our corpus is composed of three parts, each presenting different features with respect to linguistic complexity: (1) Child-directed utterances of the publicly available North American and United Kingdom portions of the CHILDES database~\cite{macwhinney2000childes}; (2) movie and TV series subtitles from the OpenSubtitle corpus~\cite{lison2016opensubtitles2016}, filtered by content-rating label (G for movies and TV-Y, TV-Y7, TV-G for TV series), available from \textit{The Movie Database}\footnote{\url{https://www.themoviedb.org/}}; (3) a 2019 snapshot of Simple English Wikipedia\footnote{\url{https://simple.wikipedia.org/}}, an English-language edition of Wikipedia written in basic English.

These different corpora vary in size: for our experiments we randomly (with uniform probability) extract sentences from each source so that the total number of tokens approximates 3 millions (10\% are kept for validation and 10\% for testing).

\subsection{Language models}
\label{sec:lm}

For each of the considered corpora, we train a character-based LSTM on the tokenized, raw text. To do so, we slightly modify the PyTorch implementation of a vanilla LSTM.\footnote{\url{https://github.com/pytorch/examples/tree/master/word_language_model}}, adapting it to a character-based setting. We run a Bayesian optimization process~\cite{bayesoptimizer} to select the best hyperparameters for the corpus (values can be found in the supplementary material). We then produce a model every 5 epochs of training (for a total of 7 models for CHILDES, 9 models for Open Subtitles and 7 models for simple Wikipedia), as to be able to produce \textit{snapshots} of the network's abilities at different stages during training. For each of the saved models, we sample\footnote{The sampling happens as follows: a random initial letter is picked, with a probability depending on the distribution of letters at the beginning of sentences in the input data, then letters are sampled with a greedy algorithm until an \textit{end of sentence} marker is reached or the length surpasses the average sentence length of the input plus 2 standard deviations.} utterances until we reach the size of the input (the `babbling' stage).
An example of babbling is reported in Table~\ref{tab:babbling-examples}.

\begin{table*}[t]
\centering
\begin{small}
\begin{tabular}{p{0.1\textwidth}|p{0.28\textwidth}p{0.28\textwidth}p{0.28\textwidth}}
 & \textbf{CHILDES} & \textbf{opensubtitles} & \textbf{simplewiki} \\ \hline
\textbf{input} & \begin{tabular}[c]{@{}p{0.28\textwidth}@{}}you tinker tot\\ let 's see if I can make this turn here\\ that 's Jim 's business \end{tabular} & \begin{tabular}[c]{@{}p{0.28\textwidth}@{}} this is no way to treat a lion !\\ re-entry into earth 's atmosphere in 37 minutes .\\ are you worried i 'm going to try and stop you ?\end{tabular} & \begin{tabular}[c]{@{}p{0.28\textwidth}@{}}She is the rector of the National Autonomous University of Honduras ( UNAH ) since 2009 .\end{tabular} \\ \hline
\textbf{best model} & \begin{tabular}[c]{@{}p{0.28\textwidth}@{}}she 's a fire\\ if I put it down for a snack\\ oh I love you \end{tabular} & \begin{tabular}[c]{@{}p{0.28\textwidth}@{}}some of horases are here down .\\ just lost it all .\\ i said we 've ... lucky .\end{tabular} & \begin{tabular}[c]{@{}p{0.28\textwidth}@{}} She is a former municipality in the center of an arrondissement in the southwest of France .\end{tabular}
\end{tabular}
\caption{Examples from input text and babbling produced by the best model, for each corpus. Sentences have been sampled according to the distribution of sentence lengths in the data.}
\label{tab:babbling-examples}
\end{small}
\end{table*}

\subsection{Extracting catenae}

\begin{figure}[t]
    \centering
    \includegraphics[width=0.45\textwidth]{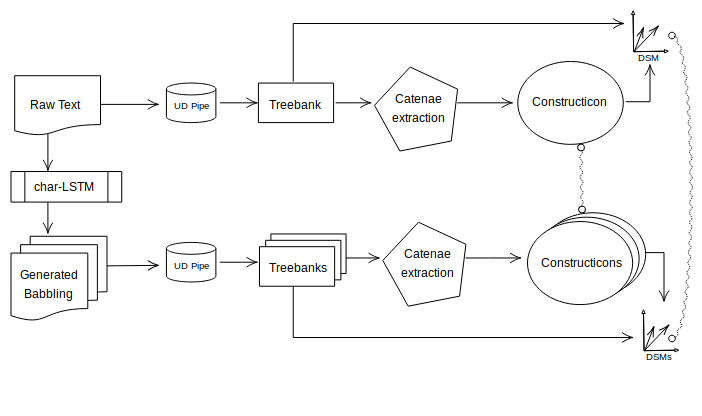}
    \caption{The  figure  depicts  the  processing  pipeline used for the experiments: raw text from corpora serves as input to the LSTM, that in turn produces raw text at  different  training  stages  (i.e.,  the babbling).   Both the  corpus  and  the  babbling  texts  are  then  processed with a NLP pipeline in order to build treebanks, from which catenae are  then  extracted. These  extracted structures form the constructicons, which are compared in the experiments described in Section~\ref{sec:q1} (dashed line). The structures in each constructicon are then represented in a Distributional Semantic Model,  through co-occurrences extracted from the respective treebanks. The distributional semantic models are then used for the experiments in \S~\ref{sec:q2} (dashed line).}
    \label{fig:pipeline}
\end{figure}

As introduced in \S~\ref{sec:framework}, the outcome of the acquisition process is a language sample $\ell_i$, that we want to compare to the input language $I_i$ or to other language samples $\ell_j$ produced at different stages of acquisition.
For the next steps, both the input text (the corpus) and the network's babbling are linguistically processed and annotated up to the syntactic level with the UDPipe toolkit~\cite{udpipe:2017} (a schema of the full processing pipeline is presented in Figure~\ref{fig:pipeline}).
Since our aim is to monitor the syntactic behaviour of the network throughout learning, we extract catenae from the input corpus and from each babbling stage. To do so, we perform a recursive depth-first visit of dependency trees (pseudocode is provided in the  supplementary material). That is, if the node $A$ is a leaf, then the only possible catena is the one containing $A$ itself; otherwise, all catenae rooted in $A$ are formed by $A$ plus a (eventually empty) combination of catenae rooted in its children nodes.

With this procedure, we extract catenae from sentences (with length between 1 and 25). For efficiency reasons, we exclude catenae longer than 5 elements. Many  structures are generated, not all of which are relevant: since we see catenae as pieces of the lexicon, frequency is not the only relevant parameter and elements should be positively associated in order to be recorded as objects. 
We therefore weigh the produced structures with a multivariate version of Mutual Information (MI), based on \citet{van2011two}:
\vspace{-3mm}

\begin{footnotesize}
\begin{equation}
MI(x_1, ..., x_n) = f(x_1, ..., x_n) \log_2 \frac{p(x_1, ..., x_n)}{\prod_{i=1}^{n} p(x_i)}
\end{equation}
\end{footnotesize}

\noindent where $p(x_1, ..., x_m) = \frac{f(x_1, ..., x_m)}{\sum_{(y_1, ..., y_m)} f(y_1, ..., y_m)}$ .

Table~\ref{tab:mutual-information} shows some of the structures with highest and lowest MI: from a qualitative perspective, it is evident that the measure is able to isolate linguistically relevant patterns, such as the basic intransitive and transitive structures (\textit{@nsubj @root} and \textit{@nsubj \_VERB @obj}).

\begin{table}[t]
\begin{center}
\begin{footnotesize}
	\begin{tabular}{lll}
		\textbf{catena}                    & \textbf{frequency} & \textbf{mi} \\ \hline
		\textbf{largest mi}                &                    &             \\ \hline
		@nsubj @root                       & 294.59K            & 633.93K     \\
		\_DET \_NOUN                       & 189.97K            & 552.32K     \\
		\_VERB @obj                        & 190.72K            & 520.82K     \\
		\_PRON \_VERB                      & 271.44K            & 503.17K     \\
		@nsubj \_AUX @root                 & 129.60K            & 478.86K     \\ \hline
		\textbf{smallest mi}               &                    &             \\ \hline
		\_PRON @nsubj                      & 17.50K             & -35.54K     \\
		@root @nsubj                       & 27.61K             & -34.89K     \\
		@nsubj \_PRON                      & 11.63K             & -30.47K     \\
		\_VERB @nsubj                      & 12.79K             & -26.82K     \\
		\_AUX \_PRON                       & 15.75K             & -26.67K     \\
	\end{tabular}
	\caption{Examples of catenae extracted from CHILDES. Largest and smallest mutual information are reported, in top and bottom tier of the table respectively. Part of Speech are prefixed by ``\_'' and syntactic relations are prefixed by ``@''}
	\label{tab:mutual-information}	
\end{footnotesize}
\end{center}
\end{table}

It is important to remark that the linguistic annotation process (except for the tokenization step) and the catenae extraction processes are completely independent from the language modeling performed by the LSTM, which is only fed with raw text and is therefore completely agnostic about the linguistic categories superimposed by the parser.

\section{What do ANNs approximate?}
\label{sec:q1}

Our first analysis demonstrates that the language generated by the LSTM reproduces the distribution of the input, and that this happens well beyond the lexical level: in other words, the network has acquired statistical regularities at the level of grammatical patterns, and is able to use them productively to generate novel language fragments that adhere to the same distribution as the input. 

\begin{figure}[t]
    \centering
    \includegraphics[width=0.35\textwidth]{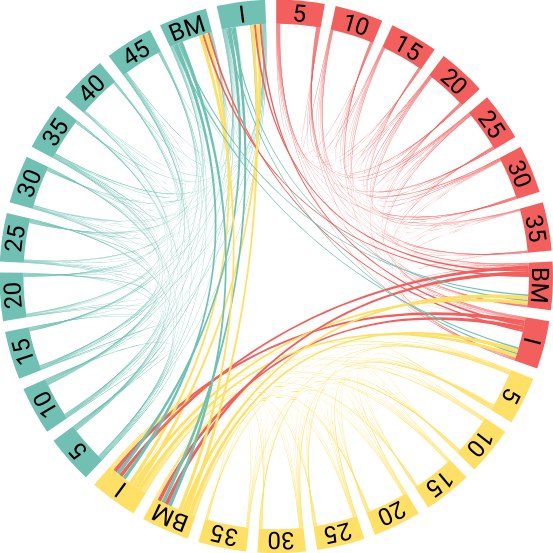}
    \caption{Correlation values (Spearman $\rho$) over top 10K catenae for each corpus (OpenSubtitles in green on the left of the plot, CHILDES in red in the top right and Simple Wikipedia in yellow at the bottom) compared to the respective babbling (at intermediate stages of learning) and the best models (BM). The thickness of the connections is inversely proportional to correlation.} 
    \label{fig:spearman-corpus}
\end{figure}

Fig.~\ref{fig:spearman-corpus} shows the extent of this approximation for various pairs: (i) $(\ell_i^c,\ell_j^c) \in \ell_{1...k}^c$ (language fragments output by a particular stage of babbling, for each corpus $c$), (ii) $(\ell_i^c, I^c), \ell_i^c \in \ell_{1...k}^c$ (fragments output by a particular stage of babbling, compared to those extracted from the respective input $c$), (iii) $(I^{c_i}, I^{c_j}), (BM^{c_i}, BM^{c_j}), (I^{c_i}, BM^{c_j})$ (fragments extracted from the input or the best babbling stage, compared among different corpora $c_i, c_j$). 
It emerges from the plot that correlations are very high within each corpus (on average, $0.935$ for CHILDES, $0.929$ for OpenSubtitles and $0.917$ for Simple Wikipedia). In particular, the correlations between the best models ($BM$) and the respective input series ($I$) show values that are among the highest, demonstrating that the network acquires structures and reproduces them with a distribution that almost perfectly matches the input. On the other hand, it is clear that different corpora show different distributions, as correlations between pairs of input series $I$ and best models show much lower values\footnote{The complete set of correlation values is reported in supplementary material}.
Overall, CHILDES scores the best correlation values, probably due to the specific features of child-directed speech, specifically its repetitiousness~\citet{clark2009first}. OpenSubtitles interestingly shows intermediate properties, sharing quite a lot of catenae with CHILDES,\footnote{The Jaccard index between CHILDES and OpenSubtitles remains above $0.5$, even when considering the top 1M catenae, while the same index computed between CHILDES and Simple Wikipedia drops to around $0.13$.} while Simple Wikipedia shows a completely different distribution.

\begin{figure}[t]
    \centering
    \includegraphics[width=0.5\textwidth]{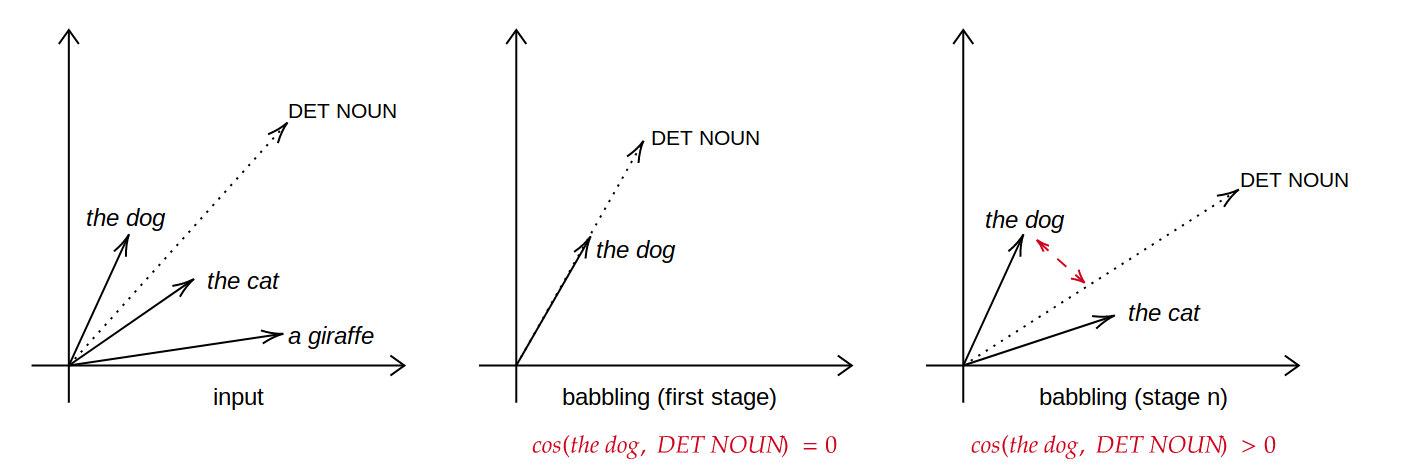}
    \caption{Let us assume that the input presents various lexicalized instances of the pattern \textit{DET NOUN} (e.g. \textit{the dog}, \textit{the cat}, \textit{a giraffe}). Our hypothesis is that the network will only be able to capture its more stereotypical instances (i.e., \textit{the dog}), and the distributions of \textit{the dog} and \textit{DET NOUN} will thus  almost perfectly overlap in the first stages of babbling (the length of vectors in the figure is just for exemplification). At later stages, the language produced by the network will show greater traits of productivity: the distribution of \textit{DET NOUN} might show that its cosine distance to \textit{the dog} has increased as it is now instantiated by two different lexicalized patterns (\textit{the dog} and \textit{the cat}) that are produced in dissimilar contexts.}
    \label{fig:distributional shift}
\end{figure}

\section{Meaning and abstraction}
\label{sec:q2}

Our second analysis relies on the idea that we can state that the network \textit{has learned some grammar} once it is able to use an acquired pattern in a productive and creative way. 
Following the basic hypothesis of CxG, stated in \S~\ref{sec:framework-rep}, we expect this generalization ability to evolve during training and the distributional properties of patterns to be in relation with the grammatical abilities of the network at various stages of learning.

Let's consider the structures $cat_1:$ \textit{the dog} and $cat_2:$ \textit{DET NOUN}. For the purpose of our analysis, we will consider $(cat_1,cat_2)$ to be a minimal pair, as \textit{the dog} can be considered a lexicalized instance of the more abstract construction \textit{DET NOUN}. Using a distributional analysis, we can capture how the contexts of $cat_1$ and $cat_2$ vary, and how this variation is associated with generalization. If their cosine similarity decreases during training, it means that their contexts become more and more dissimilar: the network produces \textit{DET NOUN} in new contexts which do not perfectly overlap with those of \textit{the dog}, indicating that the network's babbling is becoming more productive (a graphical representation is given in Figure~\ref{fig:distributional shift}). In this case, we theorize that $cat_2$ has been recognised as a partially independent pattern from $cat_1$. If, on the contrary, their cosine similarity increases, we might deduce that the network has recognized $cat_2$ as partly unnecessary: it is correcting an overgeneralization.

We restrict this analysis to the CHILDES corpus. We build distributional vector spaces for the input and each stage of babbling using the DISSECT toolkit~\cite{dinu-etal-2013-dissect}. We consider catenae composed by 2 or 3 elements as targets/contexts, and define co-occurrence as the presence of two catenae in the same sentence. Co-occurrences are weighted with PPMI and the space reduced to 300 dimensions with SVD. We then extract minimal pairs $(cat_1, cat_2)$ of catenae from the input text, where $cat_1$ is an instance of $cat_2$. For each pair, we compute their cosine similarity in all distributional spaces, and the difference in cosine between the last and first babbling (see Table~\ref{tab:pairs}).

\begin{table*}[t]
\centering
\begin{scriptsize}
\begin{tabular}{llcccccccccc}
$cat_1$ & $cat_2$ & \textbf{input} & \textbf{BM} & \textbf{5} & \textbf{10} & \textbf{15} & \textbf{20} & \textbf{25} & \textbf{30} & \textbf{35} & \textbf{distributional shift} \\ \hline
a minute & a \_NOUN & 0.28 & 0.32 & 0.71 & 0.51 & 0.44 & 0.39 & 0.38 & 0.37 & 0.34 & 0.37 \\
a minute & a @root & 0.13 & 0.19 & 0.49 & 0.37 & 0.26 & 0.20 & 0.21 & 0.22 & 0.20 & 0.30 \\
you \_VERB it & \_PRON @root @expl & 0.10 & 0.19 & 0.46 & 0.28 & 0.25 & 0.25 & 0.19 & 0.17 & 0.21 & 0.25 \\
you \_VERB you & you \_VERB @iobj & 0.28 & 0.40 & 0.68 & 0.56 & 0.47 & 0.49 & 0.39 & 0.42 & 0.43 & 0.25 \\
we can \_VERB & \_PRON can @root & 0.51 & 0.54 & 0.79 & 0.74 & 0.59 & 0.54 & 0.55 & 0.61 & 0.57 & 0.22 \\ \hline
go \_VERB @obj & \_VERB @conj @obj & 0.64 & 0.72 & 0.56 & 0.74 & 0.70 & 0.74 & 0.72 & 0.72 & 0.72 & -0.16 \\
\_AUX hungry & @cop @conj & 0.68 & 0.52 & 0.36 & 0.39 & 0.44 & 0.45 & 0.47 & 0.42 & 0.59 & -0.24 \\
can get & can @advcl & 0.55 & 0.54 & 0.24 & 0.36 & 0.45 & 0.48 & 0.43 & 0.39 & 0.52 & -0.28
\end{tabular}
\caption{Pairs of catenae $(cat_1, cat_2)$, their cosine similarity in the space obtained from CHILDES, in the space obtained from the best model (BM) and in all the intermediate models. The last column shows the difference between cosine similarity at epoch 5 and cosine similarity at epoch 35.}
\label{tab:pairs}
\end{scriptsize}
\end{table*}

We then compute average distributional shifts and cosine similarities, grouping all pairs by $cat_1$ and $cat_2$ values (for instance, we average all pairs that show abstractions of $cat_1:$ \textit{a minute}, as well as pairs that show instantiations of $cat_2:$ \textit{DET NOUN}). Some averages are shown in Table~\ref{tab:dist-shift}.

\begin{table}[t]
\begin{scriptsize}
\begin{tabular}{p{0.1\textwidth}cc|p{0.1\textwidth}cc}
$cat_1$ & \textbf{shift} & \textbf{cosine}  & $cat_2$ &  \textbf{shift} & \textbf{cosine}  \\ \hline
\tiny{@nsubj @root so} & 0.18 & 0.43 & \tiny{more @root} & 0.2 & 0.21 \\
\tiny{@nsubj only @root} & 0.18 & 0.41 & \tiny{\_AUX know @obj} & 0.19 & 0.66 \\
\tiny{what @root @obj} & 0.18 & 0.39 & \tiny{@advmod tell} & 0.17 & 0.64 \\
\tiny{what @advmod \_VERB} & 0.16 & 0.19 & \tiny{@aux know @obj} & 0.16 & 0.71 \\
\tiny{only @root} & 0.16 & 0.38 & \tiny{@advmod can \_VERB} & 0.15 & 0.76 \\
\tiny{more @root} & 0.16 & 0.23 & \tiny{know @obj} & 0.15 & 0.62 \\
\tiny{@root it @xcomp} & 0.15 & 0.61 & \tiny{a \_NOUN} & 0.13 & 0.52 \\
\tiny{@det minute} & 0.15 & 0.25 & \tiny{might @root} & 0.13 & 0.70 \\
\tiny{\_PRON only @root} & 0.15 & 0.53 & \tiny{\_PRON @root n't} & 0.12 & 0.53 \\
\tiny{\_VERB \_DET minute} & 0.15 & 0.33 & \tiny{@root that \_VERB} & 0.12 & 0.65 \\
\tiny{\_PRON @root so} & 0.14 & 0.54 & \tiny{\_VERB 'll @ccomp} & 0.12 & 0.71 \\
\tiny{\_DET minute} & 0.134 & 0.33 & \tiny{\_VERB me @obl} & 0.12 & 0.76 \\
\end{tabular}
\caption{Catenae with highest average shifts.}
\label{tab:dist-shift}
\end{scriptsize}
\end{table}

We finally split catenae in three bins based on average distributional shift and investigate the influence of input similarity over the abstraction behaviour of a construction. Our hypothesis is that catenae that underwent the highest shifts during training were those showing intermediate levels of similarities in the input distributional space. Indeed, pairs with very high input similarities are unlikely to exhibit abstraction: according to constructionist intuition, their distributional similarity means that the catena that is part of the \textit{Constructicon} is the least abstract one, and there is no need for the more abstract category. Low similarity pairs, on the other hand, may simply contain unrelated catenae. 

\begin{figure}[t]
	\centering
	\includegraphics[width=0.4\textwidth]{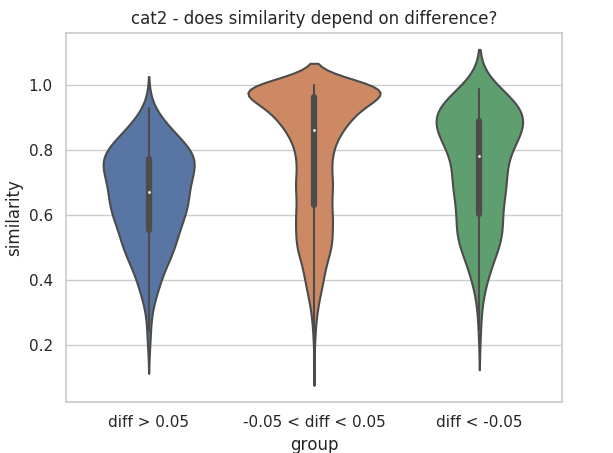}
	\caption{Distribution of average cosine similarities for the three groups of $cat_2$, showing low, intermediate and high average shifts respectively.}
	\label{fig:similarity-difference-2}
\end{figure}

To test our hypothesis, we perform a Kruskall-Wallis one-way analysis of variance test, that turn out to be significant for groupings made on both $cat_1$ and $cat_2$ lists.\footnote{$p = 6.988142426844016\text{e-}28$ for $cat_1$ and $p = 7.420868598608134\text{e-}32$ for $cat_2$} The result is confirmed by Dunn's posthoc test. We show results for the test performed on the $cat_2$ list in Table~\ref{tab:cat2-posthoc-diffs} and Figure~\ref{fig:similarity-difference-2}.

\begin{table}[t]
\begin{footnotesize}
	\begin{tabular}{l|lll}
		                         & negative    &   none   &     positive       \\ \hline
negative           &          -        &    $6.83\text{e-}06$            &   $4.57\text{e-}05$         \\
none  &    $0.000$     &  -                           &    $4.15\text{e-}29$        \\
positive            &  $0.000$       &    $4.15\text{e-}29$            &    -                     \\ 
	\end{tabular}
	\caption{Dunn posthoc test on the three groups of $c_2$, showing low ($<-0.05$), intermediate ($-0.05 < x < 0.05$) and high ($< 0.05$) average shifts respectively.}
	\label{tab:cat2-posthoc-diffs}
\end{footnotesize}
\end{table}

\section{Discussion and future work}
\label{sec:discussion}

Usage-based computational accounts have already shown to be able to explain puzzling phenomena in acquisition~\cite{freudenthal2015simulating,mccauley2019language} or to induce syntactic rules in an unsupervised manner~\cite{solan2005unsupervised}, making use of surface properties of the language signal like transitional probabilities or basic distributional analysis. However, despite being rooted in the psychological literature and yielding fundamental psycholinguistic results, the models presented in such investigations are often not comparable to studies involving neural language models, as the former are usually less flexible and less scalable to large amounts of data than the latter.

In this paper, we have reviewed relevant work concerning the assessment of grammatical abilities in neural language models and noted the lack of variety in both the input data fed to ANNs ($I$) and the theoretical framework used in analysing the output language ($\Lambda$). In line with the existing usage-based computational accounts, we have introduced a methodology to evaluate the level of productivity of an LSTM trained on limited, child-directed data, using inspirations from constructionist approaches.

We have been able to show that neural networks approximate the distribution of constructions at a quite refined level when trained over a bare 3M words from the CHILDES corpus, reproducing the distribution of grammatical patterns even when they are not fully lexicalized. The analysis in \S~\ref{sec:q1} indicates that the linguistic variety of OpenSubtitles is a potentially relevant benchmark to further investigate language acquisition, due to its similarity to the CHILDES data. In contrast, Simple Wikipedia has proved to be dissimilar to child-directed speech. This large difference should be taken into consideration when it comes to evaluating the grammatical abilities on the network: many of the studies cited in \S~\ref{sec:relwork} use models trained on Wikipedia or similar varieties, which may complicate the acquisition of generic grammatical phenomena heavily present in child-directed language. 
The analysis in \S~\ref{sec:q2} further illustrated how we can follow paths of abstraction by putting our grammar formalism in a vector space. Additional investigations are of course needed to confirm our results. In particular, we would like to target the behavior of some specific sets of structures.

Most importantly, the introduced methodology, despite being preliminary, presents a number of features that make our study fit in the usage-based theoretical framework while also using neural networks as language modeling tools, more specifically: (i) it posits no sharp distinction between lexicon and grammar: fully lexicalized, partially filled and purely syntactic patterns are all part of our constructicon and can play a similar role in production. Different items can therefore be represented compared, irrespective of their \textit{lexical} nature; (ii) it makes no assumption about the stability of the constructicon: what is relevant for productivity at the earliest stages of learning might become superfluous later on; (iii) all items are seen as form-meaning pairs (i.e., constructions by definition, as in~\citealp{goldberg2006constructions}): a novel way of modeling constructional meaning is therefore introduced and represents a promising path for future studies; (iv) distributional semantics is used both as a powerful quantitative tool and as a usage-based cognitive hypothesis, which leads us to specific assumptions about the cognitive format and origin of semantic representations~\cite{lenci2008distributional}, and seems in line with the view of constructions as ``invitations to form categories''~\cite{goldberg2019explain}.

Finally, we must account for potential biases introduced by applying dependency parsing to both input data and neural babbling: while this step is necessary to extract catenae, it introduces a non-negligible amount of noise, as the available pipelines are typically trained on different varieties than the ones considered in this study. In particular, the parser is somehow projecting its own categories, which have been acquired in a different setting and probably on a different variety, on our data. This currently limits the transferability of our results. 
Besides looking for ways to circumvent this issue, further work includes a comparison of our results with a wider choice of models.

\section*{Acknowledgements}
We thank Dr. Lucia Busso for useful discussions on an earlier version of this work and Lucio Messina for helping us in condensing results into figures and tables.
We gratefully acknowledge the support of NVIDIA Corporation with the donation of the Titan V GPU used for this research.
We would also like to thank the anonymous reviewers for their helpful suggestions and comments.

\bibliographystyle{acl_natbib}
\bibliography{main}

\begin{thebibliography}{72}
\expandafter\ifx\csname natexlab\endcsname\relax\def\natexlab#1{#1}\fi

\bibitem[{Adi et~al.(2017)Adi, Kermany, Belinkov, Lavi, and
  Goldberg}]{adi2017fine}
Yossi Adi, Einat Kermany, Yonatan Belinkov, Ofer Lavi, and Yoav Goldberg. 2017.
\newblock Fine-grained analysis of sentence embeddings using auxiliary
  prediction tasks.
\newblock In \emph{International Conference on LearningRepresentations}.

\bibitem[{Alishahi et~al.(2019)Alishahi, Chrupa{\l}a, and
  Linzen}]{alishahi2019analyzing}
Afra Alishahi, Grzegorz Chrupa{\l}a, and Tal Linzen. 2019.
\newblock Analyzing and interpreting neural networks for nlp: A report on the
  first blackboxnlp workshop.
\newblock \emph{Natural Language Engineering}, 25(4):543--557.

\bibitem[{Bacon and Regier(2019)}]{bacon2019does}
Geoff Bacon and Terry Regier. 2019.
\newblock Does bert agree? evaluating knowledge of structure dependence through
  agreement relations.
\newblock \emph{arXiv preprint arXiv:1908.09892}.

\bibitem[{Bannard et~al.(2009)Bannard, Lieven, and
  Tomasello}]{bannard2009modeling}
Colin Bannard, Elena Lieven, and Michael Tomasello. 2009.
\newblock Modeling children's early grammatical knowledge.
\newblock \emph{Proceedings of the National Academy of Sciences},
  106(41):17284--17289.

\bibitem[{Baroni(2020)}]{baroni2020linguistic}
Marco Baroni. 2020.
\newblock Linguistic generalization and compositionality in modern artificial
  neural networks.
\newblock \emph{Philosophical Transactions of the Royal Society B},
  375(1791):20190307.

\bibitem[{Bee et~al.(1969)Bee, Van~Egeren, Pytkowicz~Streissguth, Nyman, and
  Leckie}]{bee1969social}
Helen~L Bee, Lawrence~F Van~Egeren, Ann Pytkowicz~Streissguth, Barry~A Nyman,
  and Maxine~S Leckie. 1969.
\newblock Social class differences in maternal teaching strategies and speech
  patterns.
\newblock \emph{Developmental Psychology}, 1(6p1):726.

\bibitem[{Chomksy(1959)}]{chomksy1959review}
Noam Chomksy. 1959.
\newblock Review of skinner’s verbal behaviour.
\newblock \emph{Language}, 35:26--58.

\bibitem[{Chomsky(1968)}]{chomksy1968language}
Noam Chomsky. 1968.
\newblock \emph{Language and Mind}.
\newblock New York: Harcourt Brace Jovanovich.

\bibitem[{Chomsky(1981)}]{chomsky1981lectures}
Noam Chomsky. 1981.
\newblock Lectures on government and binding.
\newblock \emph{Dordrecht: Foris}.

\bibitem[{Chomsky(1995)}]{chomsky1995minimalist}
Noam Chomsky. 1995.
\newblock \emph{The minimalist program}.
\newblock MIT Press.

\bibitem[{Chowdhury and Zamparelli(2018)}]{chowdhury2018rnn}
Shammur~Absar Chowdhury and Roberto Zamparelli. 2018.
\newblock Rnn simulations of grammaticality judgments on long-distance
  dependencies.
\newblock In \emph{Proceedings of the 27th international conference on
  computational linguistics}, pages 133--144.

\bibitem[{Christiansen and Chater(2016)}]{christiansen2016creating}
Morten~H Christiansen and Nick Chater. 2016.
\newblock \emph{Creating language: Integrating evolution, acquisition, and
  processing}.
\newblock MIT Press.

\bibitem[{Clark(2009)}]{clark2009first}
Eve~V Clark. 2009.
\newblock \emph{First language acquisition}.
\newblock Cambridge University Press.

\bibitem[{Cornish et~al.(2017)Cornish, Dale, Kirby, and
  Christiansen}]{cornish2017sequence}
Hannah Cornish, Rick Dale, Simon Kirby, and Morten~H Christiansen. 2017.
\newblock Sequence memory constraints give rise to language-like structure
  through iterated learning.
\newblock \emph{PloS one}, 12(1).

\bibitem[{Crain and Pietroski(2001)}]{crain2001nature}
Stephen Crain and Paul Pietroski. 2001.
\newblock Nature, nurture and universal grammar.
\newblock \emph{Linguistics and philosophy}, 24(2):139--186.

\bibitem[{Cristia et~al.(2019)Cristia, Dupoux, Gurven, and
  Stieglitz}]{cristia2019child}
Alejandrina Cristia, Emmanuel Dupoux, Michael Gurven, and Jonathan Stieglitz.
  2019.
\newblock Child-directed speech is infrequent in a forager-farmer population: a
  time allocation study.
\newblock \emph{Child development}, 90(3):759--773.

\bibitem[{Van~de Cruys(2011)}]{van2011two}
Tim Van~de Cruys. 2011.
\newblock Two multivariate generalizations of pointwise mutual information.
\newblock In \emph{Proceedings of the Workshop on Distributional Semantics and
  Compositionality}, pages 16--20. Association for Computational Linguistics.

\bibitem[{Davis and van Schijndel(2020)}]{davis-van-schijndel-2020-recurrent}
Forrest Davis and Marten van Schijndel. 2020.
\newblock \href {https://doi.org/10.18653/v1/2020.acl-main.179} {Recurrent
  neural network language models always learn {E}nglish-like relative clause
  attachment}.
\newblock In \emph{Proceedings of the 58th Annual Meeting of the Association
  for Computational Linguistics}, pages 1979--1990, Online. Association for
  Computational Linguistics.

\bibitem[{Dinu et~al.(2013)Dinu, Pham, and Baroni}]{dinu-etal-2013-dissect}
Georgiana Dinu, Nghia~The Pham, and Marco Baroni. 2013.
\newblock \href {https://www.aclweb.org/anthology/P13-4006} {{DISSECT} -
  {DIS}tributional {SE}mantics composition toolkit}.
\newblock In \emph{Proceedings of the 51st Annual Meeting of the Association
  for Computational Linguistics: System Demonstrations}, pages 31--36, Sofia,
  Bulgaria. Association for Computational Linguistics.

\bibitem[{Dunn(2017)}]{dunn2017learnability}
Jonathan Dunn. 2017.
\newblock Learnability and falsifiability of construction grammars.
\newblock \emph{Proceedings of the Linguistic Society of America}, 2:1--1.

\bibitem[{Erk(2012)}]{erk2012vector}
Katrin Erk. 2012.
\newblock Vector space models of word meaning and phrase meaning: A survey.
\newblock \emph{Language and Linguistics Compass}, 6(10):635--653.

\bibitem[{Fillmore(1988)}]{fillmore1988mechanisms}
Charles~J Fillmore. 1988.
\newblock The mechanisms of “construction grammar”.
\newblock In \emph{Annual Meeting of the Berkeley Linguistics Society},
  volume~14, pages 35--55.

\bibitem[{Freudenthal et~al.(2015)Freudenthal, Pine, Jones, and
  Gobet}]{freudenthal2015simulating}
Daniel Freudenthal, Julian~M Pine, Gary Jones, and Fernand Gobet. 2015.
\newblock Simulating the cross-linguistic pattern of optional infinitive errors
  in children’s declaratives and wh-questions.
\newblock \emph{Cognition}, 143:61--76.

\bibitem[{Giulianelli et~al.(2018)Giulianelli, Harding, Mohnert, Hupkes, and
  Zuidema}]{giulianelli2018under}
Mario Giulianelli, Jack Harding, Florian Mohnert, Dieuwke Hupkes, and Willem
  Zuidema. 2018.
\newblock Under the hood: Using diagnostic classifiers to investigate and
  improve how language models track agreement information.
\newblock In \emph{Proceedings of the 2018 EMNLP Workshop BlackboxNLP:
  Analyzing and Interpreting Neural Networks for NLP}, pages 240--248.

\bibitem[{Goldberg(1995)}]{goldberg1995constructions}
Adele~E Goldberg. 1995.
\newblock \emph{Constructions: A construction grammar approach to argument
  structure}.
\newblock University of Chicago Press.

\bibitem[{Goldberg(2006)}]{goldberg2006constructions}
Adele~E Goldberg. 2006.
\newblock \emph{Constructions at work: The nature of generalization in
  language}.
\newblock Oxford University Press on Demand.

\bibitem[{Goldberg(2019)}]{goldberg2019explain}
Adele~E Goldberg. 2019.
\newblock \emph{Explain me this: Creativity, competition, and the partial
  productivity of constructions}.
\newblock Princeton University Press.

\bibitem[{Gulordava et~al.(2018)Gulordava, Bojanowski, Grave, Linzen, and
  Baroni}]{gulordava2018colorless}
Kristina Gulordava, Piotr Bojanowski, {\'E}douard Grave, Tal Linzen, and Marco
  Baroni. 2018.
\newblock Colorless green recurrent networks dream hierarchically.
\newblock In \emph{Proceedings of the 2018 Conference of the North American
  Chapter of the Association for Computational Linguistics: Human Language
  Technologies, Volume 1 (Long Papers)}, pages 1195--1205.

\bibitem[{Harris(1954)}]{harris1954distributional}
Zellig~S Harris. 1954.
\newblock Distributional structure.
\newblock \emph{Word}, 10(2-3):146--162.

\bibitem[{Hart and Risley(1995)}]{hart1995meaningful}
Betty Hart and Todd~R Risley. 1995.
\newblock \emph{Meaningful differences in the everyday experience of young
  American children.}
\newblock Paul H Brookes Publishing.

\bibitem[{Hauser et~al.(2002)Hauser, Chomsky, and Fitch}]{hauser2002faculty}
Marc~D Hauser, Noam Chomsky, and W~Tecumseh Fitch. 2002.
\newblock The faculty of language: what is it, who has it, and how did it
  evolve?
\newblock \emph{science}, 298(5598):1569--1579.

\bibitem[{Hewitt and Manning(2019)}]{hewitt2019structural}
John Hewitt and Christopher~D Manning. 2019.
\newblock A structural probe for finding syntax in word representations.
\newblock In \emph{Proceedings of the 2019 Conference of the North American
  Chapter of the Association for Computational Linguistics: Human Language
  Technologies, Volume 1 (Long and Short Papers)}, pages 4129--4138.

\bibitem[{Hochreiter and Schmidhuber(1997)}]{hochreiter1997long}
Sepp Hochreiter and J{\"u}rgen Schmidhuber. 1997.
\newblock Long short-term memory.
\newblock \emph{Neural computation}, 9(8):1735--1780.

\bibitem[{Hoffmann et~al.(2013)Hoffmann, Trousdale, and
  Jackendoff}]{hoffmann2013constructions}
Thomas Hoffmann, Graeme Trousdale, and Ray Jackendoff. 2013.
\newblock Constructions in the parallel architecture.

\bibitem[{Hu et~al.(2020)Hu, Gauthier, Qian, Wilcox, and
  Levy}]{hu2020systematic}
Jennifer Hu, Jon Gauthier, Peng Qian, Ethan Wilcox, and Roger~P Levy. 2020.
\newblock A systematic assessment of syntactic generalization in neural
  language models.
\newblock \emph{arXiv preprint arXiv:2005.03692}.

\bibitem[{Jawahar et~al.(2019)Jawahar, Sagot, and Seddah}]{jawahar2019does}
Ganesh Jawahar, Beno{\^\i}t Sagot, and Djam{\'e} Seddah. 2019.
\newblock What does bert learn about the structure of language?
\newblock In \emph{Proceedings of the 57th Annual Meeting of the Association
  for Computational Linguistics}, pages 3651--3657.

\bibitem[{Kay and Fillmore(1999)}]{kay1999grammatical}
Paul Kay and Charles~J Fillmore. 1999.
\newblock Grammatical constructions and linguistic generalizations: the what's
  x doing y? construction.
\newblock \emph{Language}, pages 1--33.

\bibitem[{Kuncoro et~al.(2018)Kuncoro, Dyer, Hale, Yogatama, Clark, and
  Blunsom}]{kuncoro-etal-2018-lstms}
Adhiguna Kuncoro, Chris Dyer, John Hale, Dani Yogatama, Stephen Clark, and Phil
  Blunsom. 2018.
\newblock \href {https://doi.org/10.18653/v1/P18-1132} {{LSTM}s can learn
  syntax-sensitive dependencies well, but modeling structure makes them
  better}.
\newblock In \emph{Proceedings of the 56th Annual Meeting of the Association
  for Computational Linguistics (Volume 1: Long Papers)}, pages 1426--1436,
  Melbourne, Australia. Association for Computational Linguistics.

\bibitem[{Lakretz et~al.(2019)Lakretz, Unit, Kruszewski, Desbordes, Hupkes,
  Dehaene, and Baroni}]{lakretz2019emergence}
Yair Lakretz, Cognitive~Neuroimaging Unit, German Kruszewski, Theo Desbordes,
  Dieuwke Hupkes, Stanislas Dehaene, and Marco Baroni. 2019.
\newblock The emergence of number and syntax units in lstm language models.
\newblock In \emph{Proceedings of NAACL-HLT}, pages 11--20.

\bibitem[{Lenci(2008)}]{lenci2008distributional}
Alessandro Lenci. 2008.
\newblock Distributional semantics in linguistic and cognitive research.
\newblock \emph{Italian journal of linguistics}, 20(1):1--31.

\bibitem[{Lenci(2018)}]{lenci2018distributional}
Alessandro Lenci. 2018.
\newblock Distributional models of word meaning.
\newblock \emph{Annual review of Linguistics}, 4:151--171.

\bibitem[{Lepori et~al.(2020)Lepori, Linzen, and
  McCoy}]{lepori2020representations}
Michael~A Lepori, Tal Linzen, and R~Thomas McCoy. 2020.
\newblock Representations of syntax [mask] useful: Effects of constituency and
  dependency structure in recursive lstms.
\newblock \emph{arXiv preprint arXiv:2005.00019}.

\bibitem[{Lewis and Elman(2001)}]{lewis2001learnability}
John~D Lewis and Jeffrey~L Elman. 2001.
\newblock Learnability and the statistical structure of language: Poverty of
  stimulus arguments revisited.
\newblock In \emph{Proceedings of the 26th annual Boston University conference
  on language development}, volume~1, pages 359--370. Citeseer.

\bibitem[{Lewkowicz et~al.(2018)Lewkowicz, Schmuckler, and
  Mangalindan}]{lewkowicz2018learning}
David~J Lewkowicz, Mark~A Schmuckler, and Diane~MJ Mangalindan. 2018.
\newblock Learning of hierarchical serial patterns emerges in infancy.
\newblock \emph{Developmental psychobiology}, 60(3):243--255.

\bibitem[{Lin et~al.(2019)Lin, Tan, and Frank}]{lin2019open}
Yongjie Lin, Yi~Chern Tan, and Robert Frank. 2019.
\newblock Open sesame: Getting inside bert’s linguistic knowledge.
\newblock In \emph{Proceedings of the 2019 ACL Workshop BlackboxNLP: Analyzing
  and Interpreting Neural Networks for NLP}, pages 241--253.

\bibitem[{Linzen and Baroni(2020)}]{linzen2020syntactic}
Tal Linzen and Marco Baroni. 2020.
\newblock Syntactic structure from deep learning.
\newblock \emph{arXiv preprint arXiv:2004.10827}.

\bibitem[{Linzen et~al.(2016)Linzen, Dupoux, and
  Goldberg}]{linzen2016assessing}
Tal Linzen, Emmanuel Dupoux, and Yoav Goldberg. 2016.
\newblock Assessing the ability of lstms to learn syntax-sensitive
  dependencies.
\newblock \emph{Transactions of the Association for Computational Linguistics},
  4:521--535.

\bibitem[{Lison and Tiedemann(2016)}]{lison2016opensubtitles2016}
Pierre Lison and J{\"o}rg Tiedemann. 2016.
\newblock Opensubtitles2016: Extracting large parallel corpora from movie and
  tv subtitles.

\bibitem[{Lust(1999)}]{lust1999universal}
Barbara Lust. 1999.
\newblock Universal grammar: The strong continuity hypothesis in first language
  acquisition.
\newblock \emph{Handbook of Child Language Acquisition}.

\bibitem[{MacWhinney(2000)}]{macwhinney2000childes}
Brian MacWhinney. 2000.
\newblock \emph{The CHILDES Project: Tools for analyzing talk. Third Edition.}
\newblock Lawrence Erlbaum Associates.

\bibitem[{Marvin and Linzen(2018)}]{marvin2018targeted}
Rebecca Marvin and Tal Linzen. 2018.
\newblock Targeted syntactic evaluation of language models.
\newblock In \emph{Proceedings of the 2018 Conference on Empirical Methods in
  Natural Language Processing}, pages 1192--1202.

\bibitem[{Matthews and Bannard(2010)}]{matthews2010children}
Danielle Matthews and Colin Bannard. 2010.
\newblock Children’s production of unfamiliar word sequences is predicted by
  positional variability and latent classes in a large sample of child-directed
  speech.
\newblock \emph{Cognitive science}, 34(3):465--488.

\bibitem[{McCauley and Christiansen(2019)}]{mccauley2019language}
Stewart~M McCauley and Morten~H Christiansen. 2019.
\newblock Language learning as language use: A cross-linguistic model of child
  language development.
\newblock \emph{Psychological review}, 126(1):1.

\bibitem[{McClelland(1992)}]{mcclelland1992can}
James~L McClelland. 1992.
\newblock Can connectionist models discover the structure of natural language.
\newblock \emph{Minds, Brains and Computers}, pages 168--189.

\bibitem[{McCoy et~al.(2018)McCoy, Frank, and Linzen}]{mccoy2018revisiting}
R~Thomas McCoy, Robert Frank, and Tal Linzen. 2018.
\newblock Revisiting the poverty of the stimulus: hierarchical generalization
  without a hierarchical bias in recurrent neural networks.
\newblock In \emph{Proceedings of the 40th Annual Conference of the Cognitive
  Science Society}.

\bibitem[{McCoy et~al.(2020)McCoy, Frank, and Linzen}]{mccoy2020does}
R~Thomas McCoy, Robert Frank, and Tal Linzen. 2020.
\newblock Does syntax need to grow on trees? sources of hierarchical inductive
  bias in sequence-to-sequence networks.
\newblock \emph{arXiv preprint arXiv:2001.03632}.

\bibitem[{Nivre et~al.(2020)Nivre, de~Marneffe, Ginter, Hajic, Manning,
  Pyysalo, Schuster, Tyers, and Zeman}]{nivre2020universal}
Joakim Nivre, Marie-Catherine de~Marneffe, Filip Ginter, Jan Hajic,
  Christopher~D Manning, Sampo Pyysalo, Sebastian Schuster, Francis Tyers, and
  Daniel Zeman. 2020.
\newblock Universal dependencies v2: An evergrowing multilingual treebank
  collection.
\newblock In \emph{Proceedings of The 12th Language Resources and Evaluation
  Conference}, pages 4034--4043.

\bibitem[{Nogueira(2014--)}]{bayesoptimizer}
Fernando Nogueira. 2014--.
\newblock \href {https://github.com/fmfn/BayesianOptimization} {{Bayesian
  Optimization}: Open source constrained global optimization tool for
  {Python}}.

\bibitem[{Osborne(2006)}]{osborne2006beyond}
Timothy Osborne. 2006.
\newblock Beyond the constituent-a dependency grammar analysis of chains.
\newblock \emph{Folia Linguistica}, 39(3-4):251--297.

\bibitem[{Osborne and Gro{\ss}(2012)}]{osborne2012constructions}
Timothy Osborne and Thomas Gro{\ss}. 2012.
\newblock Constructions are catenae: Construction grammar meets dependency
  grammar.

\bibitem[{Osborne et~al.(2012)Osborne, Putnam, and
  Gro{\ss}}]{osborne2012catenae}
Timothy Osborne, Michael Putnam, and Thomas Gro{\ss}. 2012.
\newblock Catenae: Introducing a novel unit of syntactic analysis.
\newblock \emph{Syntax}, 15(4):354--396.

\bibitem[{Osborne(2018)}]{osborne2018tests}
Timothy~J Osborne. 2018.
\newblock Tests for constituents: What they really reveal about the nature of
  syntactic structure.
\newblock \emph{Language Under Discussion}, 5(1):1--41.

\bibitem[{Rambelli et~al.(2019)Rambelli, Chersoni, Blache, Huang, and
  Lenci}]{rambelli2019distributional}
Giulia Rambelli, Emmanuele Chersoni, Philippe Blache, Chu-Ren Huang, and
  Alessandro Lenci. 2019.
\newblock Distributional semantics meets construction grammar. towards a
  unified usage-based model of grammar and meaning.
\newblock In \emph{Proceedings of the First International Workshop on Designing
  Meaning Representations}, pages 110--120.

\bibitem[{Ravfogel et~al.(2018)Ravfogel, Goldberg, and Tyers}]{ravfogel2018can}
Shauli Ravfogel, Yoav Goldberg, and Francis Tyers. 2018.
\newblock Can lstm learn to capture agreement? the case of basque.
\newblock In \emph{Proceedings of the 2018 EMNLP Workshop BlackboxNLP:
  Analyzing and Interpreting Neural Networks for NLP}, pages 98--107.

\bibitem[{Shen et~al.(2018)Shen, Tan, Sordoni, and Courville}]{shen2018ordered}
Yikang Shen, Shawn Tan, Alessandro Sordoni, and Aaron Courville. 2018.
\newblock Ordered neurons: Integrating tree structures into recurrent neural
  networks.
\newblock In \emph{International Conference on Learning Representations}.

\bibitem[{Solan et~al.(2005)Solan, Horn, Ruppin, and
  Edelman}]{solan2005unsupervised}
Zach Solan, David Horn, Eytan Ruppin, and Shimon Edelman. 2005.
\newblock Unsupervised learning of natural languages.
\newblock \emph{Proceedings of the National Academy of Sciences},
  102(33):11629--11634.

\bibitem[{Straka and Strakov\'{a}(2017)}]{udpipe:2017}
Milan Straka and Jana Strakov\'{a}. 2017.
\newblock \href {http://www.aclweb.org/anthology/K/K17/K17-3009.pdf}
  {Tokenizing, pos tagging, lemmatizing and parsing ud 2.0 with udpipe}.
\newblock In \emph{Proceedings of the CoNLL 2017 Shared Task: Multilingual
  Parsing from Raw Text to Universal Dependencies}, pages 88--99, Vancouver,
  Canada. Association for Computational Linguistics.

\bibitem[{Tenney et~al.(2019)Tenney, Xia, Chen, Wang, Poliak, McCoy, Kim,
  Van~Durme, Bowman, Das et~al.}]{tenney2019you}
Ian Tenney, Patrick Xia, Berlin Chen, Alex Wang, Adam Poliak, R~Thomas McCoy,
  Najoung Kim, Benjamin Van~Durme, Samuel Bowman, Dipanjan Das, et~al. 2019.
\newblock What do you learn from context? probing for sentence structure in
  contextualized word representations.
\newblock In \emph{7th International Conference on Learning Representations,
  ICLR 2019}.

\bibitem[{Tomasello(2003)}]{tomasello2003constructing}
Michael Tomasello. 2003.
\newblock \emph{Constructing a language: A usage-based theory of language
  acquisition}.
\newblock Harvard University Press.

\bibitem[{Tran et~al.(2018)Tran, Bisazza, and Monz}]{tran2018importance}
Ke~M Tran, Arianna Bisazza, and Christof Monz. 2018.
\newblock The importance of being recurrent for modeling hierarchical
  structure.
\newblock In \emph{Proceedings of the 2018 Conference on Empirical Methods in
  Natural Language Processing}, pages 4731--4736.

\bibitem[{Warstadt et~al.(2019)Warstadt, Parrish, Liu, Mohananey, Peng, Wang,
  and Bowman}]{warstadt2019blimp}
Alex Warstadt, Alicia Parrish, Haokun Liu, Anhad Mohananey, Wei Peng, Sheng-Fu
  Wang, and Samuel~R Bowman. 2019.
\newblock Blimp: A benchmark of linguistic minimal pairs for english.
\newblock \emph{arXiv preprint arXiv:1912.00582}.

\bibitem[{Wilcox et~al.(2018)Wilcox, Levy, Morita, and Futrell}]{wilcox2018rnn}
Ethan Wilcox, Roger Levy, Takashi Morita, and Richard Futrell. 2018.
\newblock What do rnn language models learn about filler--gap dependencies?
\newblock In \emph{Proceedings of the 2018 EMNLP Workshop BlackboxNLP:
  Analyzing and Interpreting Neural Networks for NLP}, pages 211--221.

\end{thebibliography}




\end{document}